\documentclass{article}

\pdfoutput=1

\usepackage[nonatbib, preprint]{neurips_2024}

\usepackage[utf8]{inputenc} 
\usepackage[T1]{fontenc}    
\usepackage{hyperref}       
\usepackage{url}            
\usepackage{booktabs}       
\usepackage{amsfonts}       
\usepackage{nicefrac}       
\usepackage{microtype}      
\usepackage{xcolor}         

\usepackage{times}
\usepackage{epsfig}
\usepackage{graphicx} 
\usepackage{amsmath}
\usepackage{amsthm}
\usepackage{amssymb}
\usepackage{cancel}
\usepackage{float}

\usepackage{algorithm}
\usepackage{algorithmic}

\usepackage{tikz}
\usepackage{float}

\def\E{\mathbf{E}}

\def\I{\mathbf{I}}

\def\e{\mathbf{e}}

\def\h{\mathbf{h}}

\def\w{\mathbf{w}}
\def\x{\mathbf{x}}

\usepackage{diagbox}

\usepackage{xparse}
\newtheorem{theorem}{Theorem}
\newtheorem{lemma}         [theorem]{Lemma}

\title{UDPM: Upsampling Diffusion Probabilistic Models}

\author{
  Shady Abu-Hussein\\
  Department of Electrical Engineering\\
  Tel Aviv University\\
  \texttt{shady.abh@gmail.com} \\
  \And
  Raja Giryes\\
  Department of Electrical Engineering\\
  Tel Aviv University\\
  \texttt{raja@tauex.tau.ac.il}
}

\begin{document}

\maketitle

\begin{abstract}
Denoising Diffusion Probabilistic Models (DDPM) have recently gained significant attention. DDPMs compose a Markovian process that begins in the data domain and gradually adds noise until reaching pure white noise. DDPMs generate high-quality samples from complex data distributions by defining an inverse process and training a deep neural network to learn this mapping. However, these models are inefficient because they require many diffusion steps to produce aesthetically pleasing samples. Additionally, unlike generative adversarial networks (GANs), the latent space of diffusion models is less interpretable. In this work, we propose to generalize the denoising diffusion process into an Upsampling Diffusion Probabilistic Model (UDPM). In the forward process, we reduce the latent variable dimension through downsampling, followed by the traditional noise perturbation. As a result, the reverse process gradually denoises and upsamples the latent variable to produce a sample from the data distribution. We formalize the Markovian diffusion processes of UDPM and demonstrate its generation capabilities on the popular FFHQ, AFHQv2, and CIFAR10 datasets. UDPM generates images with as few as three network evaluations, whose overall computational cost is less than a single DDPM or EDM step, while achieving an FID score of 6.86. This surpasses current state-of-the-art efficient diffusion models that use a single denoising step for sampling. Additionally, UDPM offers an interpretable and interpolable latent space, which gives it an advantage over traditional DDPMs. Our code is available online: \url{https://github.com/shadyabh/UDPM/}
\end{abstract}

\begin{figure*}[ht!]
    \centering
    \includegraphics[width=\linewidth, bb=0 0 175 50]{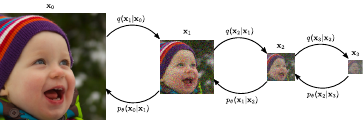}
    \caption{The Upsampling Diffusion Probabilistic Model (UDPM) scheme for 3 diffusion steps ($L=3$). In addition to the gradual noise perturbation in traditional DDPMs, UDPM also downsamples the latent variables. Accordingly, in the reverse process, UDPM denoises and upsamples the latent variables to generate images from the data distribution.}
    \label{fig:UDPM_scheme}
\end{figure*}

\section{Introduction}
In recent years, Denoising Diffusion Probabilistic Models (DDPMs) have become popular for image generation due to their ability to learn complex data distributions and generate high-fidelity images. 
These models work by starting with data samples and gradually adding noise through a Markovian process until reaching pure white noise.
This process is known as the forward diffusion process, defined by the joint distribution $q(\x_{0:L})$. The reverse diffusion process, which is used for generating new samples, is defined by the learned reverse process $p_\theta(\x_{0:L})$ using a deep neural network. This methodology allows them to achieve impressive performance in learning data distributions and sampling from them.

Although DDPMs have shown impressive results in image generation, they possess some limitations. One major limitation is that they require a large number of denoising diffusion steps to produce aesthetically pleasing samples. This can be quite intensive computationally, which makes the sampling process slow and resource-intensive. 

Additionally, the latent space of these models is not interpretable, which limits their utility for certain types of image generation tasks, such as video generation or animation, especially when used in an unconditional setting. The vast majority of works using diffusion models for editing rely on manipulating the CLIP \cite{CLIP} embeddings used with these models and not the latent space itself. 

In this work, we propose a generalized scheme of DDPMs called the Upsampling Diffusion Probabilistic Model (UDPM). In addition to the gradual noise addition in the forward diffusion process, we downsample the latent diffusion variables in order to ``dissolve'' the data information spatially, as demonstrated in Figure \ref{fig:UDPM_scheme}. We thoroughly formulate the generalized model and derive the assumptions required for obtaining a viable scheme.

Using our approach, we can sample images from CIFAR10 \cite{cifar10}, AFHQv2 \cite{choi2020stargan}, and FFHQ \cite{karras2019style} datasets, with as few as 3 UDPM steps, where the cost of all three steps together is $\sim$$30\%$ of a single regular diffusion step. 
This is less than two orders of magnitude compared to standard DDPMs: guided diffusion \cite{guidedDiff} typically requires 1000 iterations, denoising diffusion implicit models \cite{Song2022DDIM} require $250$ iterations, stable diffusion \cite{rombach2022high} requires at least 50 iterations with an additional decoder, EDM \cite{karras2022elucidating} can reduce the number of steps to 39, and other recent works \cite{lu2022dpm, lu2022dpmplus, dockhorn2022genie} can sample with 10-20 network evaluations, which is still considerably more than UDPM. Indeed, in \cite{DDGAN}, an integration of discriminative loss with diffusion models has shown great generation performance while requiring only 2 diffusion steps. Recent works \cite{zheng2022truncated, luhman2021knowledge, salimans2022progressive} have demonstrated that sampling can be performed with as few as a single denoising diffusion step while maintaining competitive generation performance. Yet, UDPM achieves better generation quality than these works on the CIFAR10 dataset \cite{cifar10} while requiring a smaller computational cost ($1/3$ of a typical diffusion step).

Because UDPM gradually reduces the dimensions of the latent variables in each step, the size of the random noise added is considerably smaller. Specifically, all the dimensions of the latent variables together are smaller than the dimensions of the original images. As a result, UDPM is much more interpretable compared to conventional DDPMs. In our experiments, we show how one may manipulate the generated images by changing the latent variables, which is similar to what has been done in Generate Adversarial Networks (GANs) \cite{karras2019style}.

Our contributions may be summarized as follows: (i) A novel efficient diffusion model for image generation that achieves
a significant improvement over current state-of-the-art methods by reducing the number and cost of diffusion steps required to generate high-quality images; (ii) achieving good interpretability and interpolability of the latent space. 

\begin{figure*}[t]
    \centering      
        \includegraphics[width=0.078\linewidth, bb=0 0 64 64]{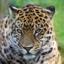}
        \includegraphics[width=0.078\linewidth, bb=0 0 64 64]{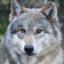}
        \includegraphics[width=0.078\linewidth, bb=0 0 64 64]{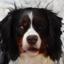}
        \includegraphics[width=0.078\linewidth, bb=0 0 64 64]{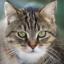}
        \includegraphics[width=0.078\linewidth, bb=0 0 64 64]{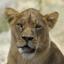}
        \includegraphics[width=0.078\linewidth, bb=0 0 64 64]{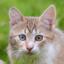}
        \includegraphics[width=0.078\linewidth, bb=0 0 64 64]{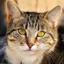}
        \includegraphics[width=0.078\linewidth, bb=0 0 64 64]{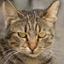}
        \includegraphics[width=0.078\linewidth, bb=0 0 64 64]{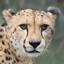}
        \includegraphics[width=0.078\linewidth, bb=0 0 64 64]{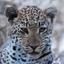}
        \includegraphics[width=0.078\linewidth, bb=0 0 64 64]{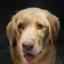}
        \includegraphics[width=0.078\linewidth, bb=0 0 64 64]{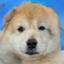}\\
        \includegraphics[width=0.078\linewidth, bb=0 0 64 64]{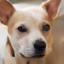}
        \includegraphics[width=0.078\linewidth, bb=0 0 64 64]{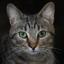}
        \includegraphics[width=0.078\linewidth, bb=0 0 64 64]{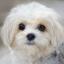}
        \includegraphics[width=0.078\linewidth, bb=0 0 64 64]{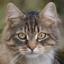}
        \includegraphics[width=0.078\linewidth, bb=0 0 64 64]{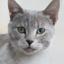}
        \includegraphics[width=0.078\linewidth, bb=0 0 64 64]{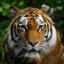}
        \includegraphics[width=0.078\linewidth, bb=0 0 64 64]{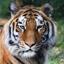}
        \includegraphics[width=0.078\linewidth, bb=0 0 64 64]{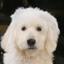}
        \includegraphics[width=0.078\linewidth, bb=0 0 64 64]{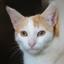}
        \includegraphics[width=0.078\linewidth, bb=0 0 64 64]{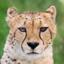}
        \includegraphics[width=0.078\linewidth, bb=0 0 64 64]{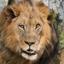}
        \includegraphics[width=0.078\linewidth, bb=0 0 64 64]{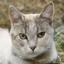}\\
        \includegraphics[width=0.078\linewidth, bb=0 0 64 64]{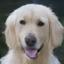}
        \includegraphics[width=0.078\linewidth, bb=0 0 64 64]{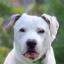}
        \includegraphics[width=0.078\linewidth, bb=0 0 64 64]{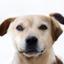}
        \includegraphics[width=0.078\linewidth, bb=0 0 64 64]{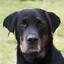}
        \includegraphics[width=0.078\linewidth, bb=0 0 64 64]{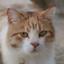}
        \includegraphics[width=0.078\linewidth, bb=0 0 64 64]{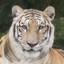}
        \includegraphics[width=0.078\linewidth, bb=0 0 64 64]{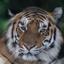}
        \includegraphics[width=0.078\linewidth, bb=0 0 64 64]{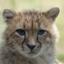}
        \includegraphics[width=0.078\linewidth, bb=0 0 64 64]{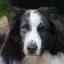}
        \includegraphics[width=0.078\linewidth, bb=0 0 64 64]{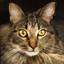}
        \includegraphics[width=0.078\linewidth, bb=0 0 64 64]{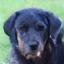}
        \includegraphics[width=0.078\linewidth, bb=0 0 64 64]{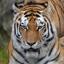}\\
        \includegraphics[width=0.078\linewidth, bb=0 0 64 64]{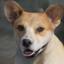}
        \includegraphics[width=0.078\linewidth, bb=0 0 64 64]{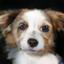}
        \includegraphics[width=0.078\linewidth, bb=0 0 64 64]{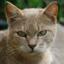}
        \includegraphics[width=0.078\linewidth, bb=0 0 64 64]{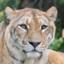}
        \includegraphics[width=0.078\linewidth, bb=0 0 64 64]{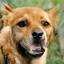}
        \includegraphics[width=0.078\linewidth, bb=0 0 64 64]{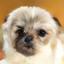}
        \includegraphics[width=0.078\linewidth, bb=0 0 64 64]{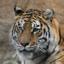}
        \includegraphics[width=0.078\linewidth, bb=0 0 64 64]{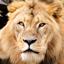}
        \includegraphics[width=0.078\linewidth, bb=0 0 64 64]{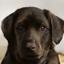}
        \includegraphics[width=0.078\linewidth, bb=0 0 64 64]{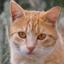}
        \includegraphics[width=0.078\linewidth, bb=0 0 64 64]{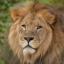}
        \includegraphics[width=0.078\linewidth, bb=0 0 64 64]{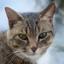}\\
    \caption{Generated $64\times 64$ images of AFHQv2 
    \cite{choi2020stargan}
    with \textbf{FID=7.10142}, produced using unconditional UDPM with only 3 steps, which are equivalent to 0.3 of a single typical $64\times 64$ diffusion step.}
    \label{fig:AFHQv264}
\end{figure*}

\section{Related Work}
\label{sec:related}

{\bf Diffusion models} are latent variable models defined through a diffusion probabilistic model. On one side we have the data $\x_0\sim q(\x_0)$ and on the other side, we have pure noise $\x_L\sim q(\x_{L})$. Both are related to each other using a Markovian diffusion process, where the forward process is defined by the joint distribution $ q(\x_{1:L}|\x_0) $ and the reverse process by $p(\x_{0:L-1}|\x_L)$. Using the Markov chain property, $ q(\x_{1:L}|\x_0)$ and $p(\x_{0:L-1}|\x_L)$ can be expressed by
\begin{equation}
    p(\x_{0:L-1}|\x_L) = \prod_{l=1}^L p(\x_{l-1}|\x_l),\label{eq:reverse_def}
\end{equation}
and 
\begin{equation}
    q(\x_{1:L}|\x_0) = \prod_{l=1}^L q(\x_l|\x_{l-1}).\label{eq:forward_def}
\end{equation}

As stated in previous literature \cite{DDPM, guidedDiff, sohl2015deep}, the common approach is to assume that the Markov chain is constructed using normal distributions defined by
\begin{equation}
    q(\x_{l}|\x_{l-1}) := \mathcal{N}(\sqrt{1 - \beta_l}\x_{l-1}, \beta_l \I),\label{eq:q_theta_def}
\end{equation}
and 
\begin{equation}
    p(\x_{l-1}|\x_{l}) := \mathcal{N}(\mu_l(\x_l), \Sigma_l(\x_l)),\label{eq:p_theta_def}
\end{equation}
where $\beta_1,\beta_2, \hdots \beta_L$ are hyperparameters that control the noise levels of the diffusion process, while $\mu_l, \Sigma_l$ denote the mean and variance of the reverse process, respectively.

By learning the reverse process $p(\x_{l-1}|\x_{l})$ using a deep neural network, one can generate samples from the data distribution by running the reverse process: starting from pure noise $\x_L\sim \mathcal{N}(0, \I)$, then progressively predicting the next step of the reverse process using a network trained to predict $\x_{l-1}$ from $\x_l$, until reaching $\x_0$. 

Over the past couple of years, this principle has shown marvellous performance in generating realistic-looking images \cite{guidedDiff, rombach2022high, DDPM, sohl2015deep, nichol2021improved,kawar2022imagic,Giannone2022Few,Sheynin2022KNN,AbuHussein2022ADIR}. However, as noted by \cite{DDPM}, the number of diffusion steps $L$ is required to be large in order for the model to produce pleasing-looking images. 
 
 Recently, many studies have utilized diffusion models for image manipulation and reconstruction tasks
 \cite{whang2022deblurring, SR3}, where a denoising network is trained to learn the prior distribution of the data. At test time, some conditioning mechanism is combined with the learned prior for solving highly challenging imaging tasks \cite{avrahami2022blended, avrahami2022blendedLatent, chung2022mr}.
Note that our novel adaptive diffusion ingredient can be incorporated into any conditional sampling scheme that is based on diffusion models.

The works in \cite{whang2022deblurring,SR3} addressed the problems of deblurring and super-resolution using diffusion models. These works try to deblur \cite{whang2022deblurring} or increase the resolution \cite{SR3} of a blurry or low-res input image. Unlike our work, their goal is not image generation, but rather image reconstruction from a given degraded image. Therefore, their trained model is significantly different from ours.

Cold diffusion \cite{Bansal2022Cold} performs diffusion steps by replacing the steps of noise addition with steps of blending with another image or other general steps. 
This approach differs significantly from ours, as their goal is to demonstrate that denoising can be replaced with other operations without developing the corresponding diffusion equations.
In our case, we formally show what operations can be used without losing the Markovian property of the forward and reverse diffusion processes, and without omitting the noise component. 

Soft diffusion \cite{daras2022soft} proposes to blur the signal before adding noise to it in the forward process. Then, for solving the reverse process they train a network to deblur and denoise the signal. However, the addition of the blur operation to the forward process prohibits explicit access to the reverse process and therefore relies solely on the network to predict the clean sharp sample. In contrast, in our method, the reverse process is explicitly accessible and can be sampled easily.

Another effort \cite{guth2022wavelet,kadkhodaie2023learning} suggests replacing the denoising steps by learning the wavelet coefficients of the high frequencies. They show that this can reduce the number of diffusion steps. Our work differs from theirs in the fact that we rely on upsampling with an additive noise step. We also employ advanced loss functions and present state-of-the-art generation results. This is in addition to our interpretable latent space, a component missing from many of the recent diffusion models.

Many works tried to accelerate the sampling procedure of the denoising diffusion model \cite{Song2022DDIM, karras2022elucidating, lu2022dpm, lu2022dpmplus, dockhorn2022genie}. However, they only focus on reducing the number of sampling steps, while ignoring the diffusion structure itself. By contrast, in this work we propose to degrade the signal not only over the noise domain, but also in the spatial domain, thereby ``dissolving'' the signal much faster.

\section{Method}

Traditional denoising diffusion models assume that the probabilistic Markov process is defined by \eqref{eq:q_theta_def} and \eqref{eq:p_theta_def}. These equations construct forward and backward processes that progress by adding and removing noise, respectively. 
In this work, we generalize this scheme by adding a degradation element to the forward process. Specifically, we downsample the spatial dimension of the latent variable and upsample it when reversing the process. 

\subsection{Upsampling Diffusion Probabilistic Model (UDPM)}\label{sec:UDPM}

We begin by redefining the marginal distributions $q(\x_{l}|\x_{l-1})$ and $p(\x_{l-1}|\x_{l})$ of the forward and reverse processes:
\begin{equation}
    q(\x_{l}|\x_{l-1}) :=  \mathcal{N}(\alpha_l \mathcal{H}\x_{l-1}, \sigma_l^2 \I),\label{eq:q_theta_def_UDPM}
\end{equation}
and 
\begin{equation}
    p(\x_{l-1}|\x_{l}) := \mathcal{N}(\mu(\x_l; l), \Sigma_l), \label{eq:p_theta_def_UDPM}
\end{equation}

where in contrast to previous diffusion models that used $\mathcal{H}=I$, in this work we define the operator $\mathcal{H}$ as a downsampling operator, defined by applying a blur filter $\mathcal{W}$ followed by subsampling with stride $\gamma$. As a result, the forward diffusion process decreases the variables' dimensions in addition to the increased noise levels.

In diffusion models, the goal is to match the joint distributions $p_\theta(\x_{1:L}|\x_0)$ (learned) and $q(\x_{1:L}|\x_0)$ under some statistical distance. One particular choice is the Kullback-Leibler (KL) divergence, which we adopt here.
Formally,
\begin{align}
    D_\text{KL}(q(\x_{1:L}|\x_0)||p_\theta(\x_{1:L}|\x_0)) := \E_q\left[ \log \frac{q(\x_{1:L}|\x_0)}{p_\theta(\x_{1:L}|\x_0)}\right] = \log p_\theta (\x_0)\underbrace{-\E_q \left[ \frac{p_\theta(\x_{0:L})}{q(\x_{1:L}|\x_0)}\right]}_{\text{ELBO}}.
    \label{eq:KL_div}
\end{align}
Thus, one can minimize the KL-divergence between $p_\theta(\x_{1:L}|\x_0)$ and $q(\x_{1:L}|\x_0)$ by minimizing the Evidence Lower Bound (ELBO). As we show in Appendix~\ref{appndx:ELBO}, this is equivalent to
\begin{align}
    \E_q \left[-\log \frac{p_\theta (\x_{0:L})}{q(\x_{1:L}|\x_0)}\right]
    &= \E_q[D_\text{KL}(p(\x_L)||q(\x_L|\x_0)) \nonumber\\
    &+ \sum_{l=2}^L D_\text{KL}(p_\theta(\x_{l-1}|\x_l)||q(\x_{l-1}|\x_l, \x_0)
    - \log p_\theta(\x_1|\x_0)].
    \label{eq:ELBO}
\end{align}
The right-hand side of \eqref{eq:ELBO} can be then minimized stochastically w.r.t. $\theta$ using gradient descent, where at each step a random $l$ is chosen and a single term of \eqref{eq:ELBO} is optimized.

In order to be able to use \eqref{eq:ELBO} for training $p_\theta(\cdot)$, one needs explicit access to $q(\x_{l-1}| \x_l, \x_0)$, for which we need to obtain $q(\x_l|\x_0)$ first. Then, using Bayes' theorem, we can derive $q(\x_{l-1}| \x_l, \x_0)$.
To do so, we first present Lemma \ref{lemma:sampled_gauss} (the proof is in Appendix \ref{appndx:lemma_proof})
\begin{lemma}\label{lemma:sampled_gauss}
Let $\e \stackrel{iid}{\sim} \mathcal{N}(0, \I) \in \mathbb{R}^N$ and $\mathcal{H} = \mathcal{S}_\gamma \mathcal{W}$, where $\mathcal{S}_\gamma$ is a subsampling operator with stride $\gamma$ and $\mathcal{W}$ is a blur operator with blur kernel $\w$. Then, if the support of $\w$ is at most $\gamma$, we have $\mathcal{H}\e \stackrel{iid}{\sim} \mathcal{N}(0, \|\w\|_2^2 \I)$.
\end{lemma}

If Lemma \ref{lemma:sampled_gauss} holds, then by assuming that $\|\w\|_2^2 = 1$, we get the following result (see Appendix \ref{appndx:q_xt_given_x0})
\begin{equation}
\label{eq:xt_given_x0}
    q(\x_l| \x_0) = \mathcal{N}(\bar{\alpha}_l \mathcal{H}^l \x_0, \tilde{\sigma}_l^2\I) \text{ where } \bar{\alpha}_l = \prod_{k=0}^l \alpha_k \text{, and }\tilde{\sigma}_l=\bar{\alpha}_l^2\sum_{k=1}^l \frac{\sigma_k^2}{\bar{\alpha}_l^2}.
\end{equation}
Using \eqref{eq:xt_given_x0}, we can obtain $\x_t$ from $\x_0$ simply by applying $\mathcal{H}$ $l$-times on $\x_0$, followed by the addition of white Gaussian noise with standard deviation $\tilde{\sigma}_l$.

Given $q(\x_l| \x_0)$, we can use Bayes' theorem and utilize the Markov chain property to get
\begin{equation*}
    q(\x_{l-1}|\x_l, \x_0) = \frac{q(\x_l|\x_{l-1})q(\x_{l-1}|\x_0)}{q(\x_l|\x_0)},
\end{equation*}
which, as shown in Appendix~\ref{appndx:q_xt1_given_xt_x0}, is of the following form
\begin{equation}
\label{eq:q_xt1_given_xt_x0}
    q(\x_{l-1}|\x_l, \x_0) = \mathcal{N}(\mu(\x_l, \x_0, l), \Sigma_l),\end{equation}
where 
\begin{align}
\label{eq:sigma_t}
    \Sigma_l = \left( \frac{\alpha_l^2}{\sigma^2_l}\mathcal{H}^T \mathcal{H} + \frac{1}{\tilde{\sigma}_{l-1}} \I \right)^{-1},
\end{align}
and 
\begin{align}
\label{eq:mu_t}
    \mu(\x_l, \x_0, l) = \Sigma_l \left( \frac{\alpha_l}{\sigma_l^2}\mathcal{H}^T \x_l + \frac{\bar{\alpha}_{l-1}}{\tilde{\sigma}^2_{l-1}} \mathcal{H}^{l-1} \x_0 \right).
\end{align}

\begin{figure*}[t]
    \centering      
    \includegraphics[width=0.078\linewidth, bb=0 0 64 64]{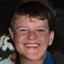}
    \includegraphics[width=0.078\linewidth, bb=0 0 64 64]{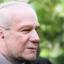}
    \includegraphics[width=0.078\linewidth, bb=0 0 64 64]{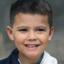}
    \includegraphics[width=0.078\linewidth, bb=0 0 64 64]{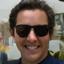}
    \includegraphics[width=0.078\linewidth, bb=0 0 64 64]{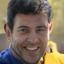}
    \includegraphics[width=0.078\linewidth, bb=0 0 64 64]{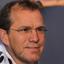}
    \includegraphics[width=0.078\linewidth, bb=0 0 64 64]{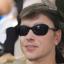}
    \includegraphics[width=0.078\linewidth, bb=0 0 64 64]{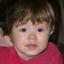}
    \includegraphics[width=0.078\linewidth, bb=0 0 64 64]{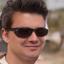}
    \includegraphics[width=0.078\linewidth, bb=0 0 64 64]{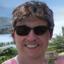}
    \includegraphics[width=0.078\linewidth, bb=0 0 64 64]{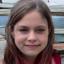}
    \includegraphics[width=0.078\linewidth, bb=0 0 64 64]{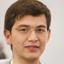}\\
    \includegraphics[width=0.078\linewidth, bb=0 0 64 64]{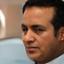}
    \includegraphics[width=0.078\linewidth, bb=0 0 64 64]{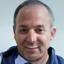}
    \includegraphics[width=0.078\linewidth, bb=0 0 64 64]{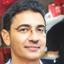}
    \includegraphics[width=0.078\linewidth, bb=0 0 64 64]{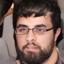}
    \includegraphics[width=0.078\linewidth, bb=0 0 64 64]{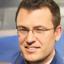}
    \includegraphics[width=0.078\linewidth, bb=0 0 64 64]{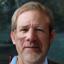}
    \includegraphics[width=0.078\linewidth, bb=0 0 64 64]{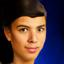}
    \includegraphics[width=0.078\linewidth, bb=0 0 64 64]{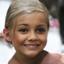}
    \includegraphics[width=0.078\linewidth, bb=0 0 64 64]{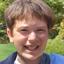}
    \includegraphics[width=0.078\linewidth, bb=0 0 64 64]{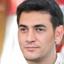}
    \includegraphics[width=0.078\linewidth, bb=0 0 64 64]{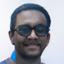}
    \includegraphics[width=0.078\linewidth, bb=0 0 64 64]{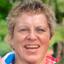}\\
    \includegraphics[width=0.078\linewidth, bb=0 0 64 64]{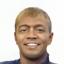}
    \includegraphics[width=0.078\linewidth, bb=0 0 64 64]{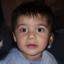}
    \includegraphics[width=0.078\linewidth, bb=0 0 64 64]{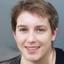}
    \includegraphics[width=0.078\linewidth, bb=0 0 64 64]{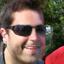}
    \includegraphics[width=0.078\linewidth, bb=0 0 64 64]{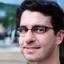}
    \includegraphics[width=0.078\linewidth, bb=0 0 64 64]{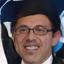}
    \includegraphics[width=0.078\linewidth, bb=0 0 64 64]{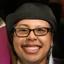}
    \includegraphics[width=0.078\linewidth, bb=0 0 64 64]{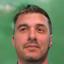}
    \includegraphics[width=0.078\linewidth, bb=0 0 64 64]{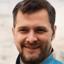}
    \includegraphics[width=0.078\linewidth, bb=0 0 64 64]{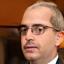}
    \includegraphics[width=0.078\linewidth, bb=0 0 64 64]{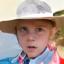}
    \includegraphics[width=0.078\linewidth, bb=0 0 64 64]{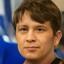}\\
    \includegraphics[width=0.078\linewidth, bb=0 0 64 64]{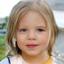}
    \includegraphics[width=0.078\linewidth, bb=0 0 64 64]{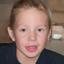}
    \includegraphics[width=0.078\linewidth, bb=0 0 64 64]{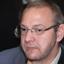}
    \includegraphics[width=0.078\linewidth, bb=0 0 64 64]{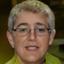}
    \includegraphics[width=0.078\linewidth, bb=0 0 64 64]{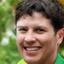}
    \includegraphics[width=0.078\linewidth, bb=0 0 64 64]{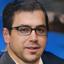}
    \includegraphics[width=0.078\linewidth, bb=0 0 64 64]{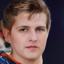}
    \includegraphics[width=0.078\linewidth, bb=0 0 64 64]{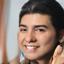}
    \includegraphics[width=0.078\linewidth, bb=0 0 64 64]{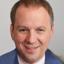}
    \includegraphics[width=0.078\linewidth, bb=0 0 64 64]{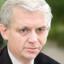}
    \includegraphics[width=0.078\linewidth, bb=0 0 64 64]{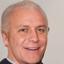}
    \includegraphics[width=0.078\linewidth, bb=0 0 64 64]{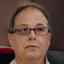}\\
    \caption{Generated $64\times 64$ images of FFHQ
    with \textbf{FID=7.41065}, produced using unconditional UDPM with only 3 steps, which are equivalent to 0.3 of a single typical $64\times64$ diffusion step.}
    \label{fig:FFHQ64}
\end{figure*}

Although expression \eqref{eq:sigma_t} seems to be implacable, in practice it can be implemented efficiently using the Discrete Fourier Transform and the poly-phase filtering identity used in \cite{corrFlt, chan2016plug}, where $\mathcal{H}^T \mathcal{H}$ is equivalent to a convolution between $\w$ and its flipped version, followed by subsampling with stride $\gamma$. More details are provided in Appendix \ref{appndx:Sigma_mu_derivation}. 

Following \eqref{eq:q_xt1_given_xt_x0}, the true posterior $q(\x_{l-1}|\x_l , \x_0)$ in this new setup is a Gaussian distribution with parameters $(\mu(\x_l, \x_0, l), \Sigma_l)$. Therefore, one may assume that $p_\theta(\cdot)$ is also Gaussian with parameters $(\mu_\theta, \Sigma_l)$, where $\mu_\theta$ is parameterized by a deep neural network with learned parameters $\theta$. 

For normal distributions, a single term of \eqref{eq:ELBO} is equivalent to
\begin{align}
    \ell^{(l)} &=  D_\text{KL}(p_\theta(\x_{l-1}|\x_l)||q(\x_{l-1}|\x_l, \x_0) \nonumber \\
    &= C_l + \frac{1}{2} (\mu_\theta - \mu_l)^T \Sigma_l^{-1} (\mu_\theta - \mu_l),\label{eq:diffusion_objective}
\end{align}
where $C_l$ is a constant value independent of $\theta$.

From our experiments and following \cite{guidedDiff, DDPM}, training $p_\theta$ to predict $\mu$ directly leads to worse results. Therefore, we train the network to predict the second term in \eqref{eq:mu_t}, i.e. to estimate $\mathcal{H}^{l-1}\x_0$ from $\x_l$. As a result, minimizing \eqref{eq:diffusion_objective} can be simplified to minimizing the following term
\begin{equation}
    \tilde{\ell}^{(l)}_\text{simple} = (f_\theta(\x_l) - \mathcal{H}^{l-1}\x_0)^T \Sigma_l^{-1} (f_\theta(\x_l) - \mathcal{H}^{l-1}\x_0),
\end{equation}
where $f_\theta(\cdot)$ is a deep neural network that upsamples its input by a scale factor of $\gamma$. 
By the definition of $\Sigma_l$, it is easy to show that it is a diagonal positive matrix, and therefore it can be dropped. As a result, one may simplify the objective to the following term
\begin{equation}
\label{eq:simple_loss}
    \ell^{(l)}_\text{simple} = \|f_\theta(\x_l) - \mathcal{H}^{l-1}\x_0\|_2^2. 
\end{equation}

Unlike denoising diffusion models, UDPM tackles a super-resolution task at each reverse step. 
Consequently, relying solely on $\ell_{\text{simple}}$ for the training of $f_\theta(\cdot)$ produces softer images. 
To address this, we propose to incorporate two additional regularization terms, following \cite{wang2021real}. 
The first term is a perceptual loss \cite{zhang2018unreasonable}, denoted by $\ell_\text{per}$, which aligns the VGG features of $\mathcal{H}^{l-1}\x_0$ and $f_\theta(\x_l)$. The second term is an adversarial loss denoted as $\ell_\text{adv}$, which is similar to the one used in \cite{wang2021real, ledig2017photo}. Both are used to ensure sharp detailed results.

\begin{figure*}[!ht]
    \centering
    \includegraphics[width=\linewidth, bb=0 0 300 120]{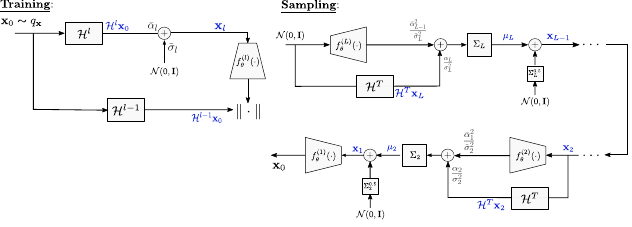}
    \caption{The training and sampling procedures of UDPM. During the \textit{training} phase, an image $\x_0$ is randomly selected from the dataset $\{\x_0\} \sim q(\x_0)$. It is then degraded using \eqref{eq:xt_given_x0} to obtain a downsampled noisy version $\x_l$, which is then plugged into a deep neural network $f_\theta^{(l)}(\cdot)$. The network is trained to predict $\mathcal{H}^{l-1}\x_0$. In the \textit{sampling} phase, a pure white Gaussian noise $\x_L \sim \mathcal{N}(0,\I)$ is generated. This noise is passed through the network $f_\theta^{(L)}(\cdot)$ to estimate $\mathcal{H}^{L-1}\x_0$. The estimated $\mathcal{H}^{L-1}\x_0$ is used to compute $\mu_L$ through \eqref{eq:mu_t}, with $\Sigma_L$ obtained from \eqref{eq:sigma_t}. Afterwards, $\x_{L-1}$ is drawn from $\mathcal{N}(\mu_L, \Sigma_L)$  using the technique described in Appendix \ref{appndx:sampling_posterior}. By repeating this procedure for $L$ iterations, the final sample $\x_0$ is obtained.}\label{fig:UDPM_train_sample_scheme}
\end{figure*}

Algorithm \ref{alg:training} and Figure \ref{fig:UDPM_train_sample_scheme} provide an overview of the UDPM training scheme.

\begin{figure*}[t]
    \centering
    \begin{minipage}[t]{0.49\textwidth}
        \begin{algorithm}[H]
        \small
            \caption{UDPM training algorithm}
            \label{alg:training}
            \begin{algorithmic}[1]
                \REQUIRE $f_\theta(\cdot), L, q(\x), D_\phi(\cdot)$
                \WHILE{Not converged}
                    \STATE $\x_0 \sim q(\x)$
                    \STATE $l \in \{1,2,\hdots, L\}$
                    \STATE $\e \sim \mathcal{N}(0, I)$
                    \STATE $\x_l = \bar{\alpha}_l \mathcal{H}^l\x_0 + \tilde{\sigma}_l \e$
                    \STATE $\ell = \lambda_\text{fid}^{(l)}\ell_\text{simple} + \lambda_\text{per}^{(l)}\ell_\text{per} + \lambda_\text{adv}^{(l)}\ell_\text{adv}$ 
                    \STATE ADAM step on $\theta$
                    \STATE Adversarial ADAM step on $\phi$
                \ENDWHILE
                \STATE \textbf{return} $f_\theta(\cdot)$
            \end{algorithmic}   
        \end{algorithm}
    \end{minipage}\hfill
    \begin{minipage}[t]{0.49\textwidth}
        \begin{algorithm}[H]
        \small
            \caption{UDPM sampling algorithm}
            \label{alg:sampling}
            \begin{algorithmic}[1]
                \REQUIRE $f_\theta(\cdot), L$
                \STATE $\x_L \sim \mathcal{N}(0, I)$
                \FORALL{$l=L,\hdots, 1$}
                    \STATE $\Sigma = \left( \frac{\alpha_l^2}{\sigma_l^2}\mathcal{H}^T \mathcal{H} + \frac{1}{\tilde{\sigma}_{l-1}^2} \I \right)^{-1}$
                    \STATE $\mu_\theta = \Sigma \left[ \frac{\alpha_l}{\sigma_l^2}\mathcal{H}^T \x_l + \frac{\bar{\alpha}_{l-1}}{\tilde{\sigma}_{l-1}^2} f^{(l)}_\theta(\x_l) \right]$
                    \STATE $\x_{l-1} \sim \mathcal{N}(\mu_\theta, \Sigma)$
                \ENDFOR
                \STATE \textbf{return} $\x_0$
            \end{algorithmic}
        \end{algorithm}
    \end{minipage}
\end{figure*}

\subsection{Image Generation using UDPM} \label{sec:generation_using_UDPM}
The UDPM scheme presented in the previous section introduces a new approach for capturing the true implicit data distribution $q(\x_0)$ of a given dataset. Given that we trained UDPM, the remaining question is how it can be utilized for sampling from $q_(\x_0)$. 

Given a deep neural network $f_\theta$ trained to predict $\mathcal{H}^{l-1}\x_0$ from $\x_l$, we start with a pure Gaussian noise sample $\x_L\sim \mathcal{N}(0, I)$. Subsequently, by substituting $f_\theta(\x_l)$ into $\mathcal{H}^{l-1}\x_0$ in \eqref{eq:mu_t}, we get an estimate of $\mu_{L}$. Then, the next reverse diffusion step $\x_{L-1}$ can be obtained by sampling from $\mathcal{N}(\mu_L, \Sigma_L)$. By iteratively repeating these steps $L$ times, we can acquire a sample $\x_0 \sim p_\theta(\x_0)$, as outlined in Algorithm \ref{alg:sampling}.

Note that sampling the posterior requires parameterizing $\mathcal{N}(\mu_l, \Sigma_l)$ in the form of
\begin{equation*}
    \x_{l-1} = \mu_l + \Sigma_{l}^{\frac{1}{2}} \e
\end{equation*}
where $\e \sim \mathcal{N}(0, I)$ has the same dimensions as $\x_{l-1}$. This requires having access to $\Sigma^{\frac{1}{2}}_l$. Due to the structure of $\Sigma_l$, it is possible to apply it on $\e$ efficiently, as shown in Appendix \ref{appndx:sampling_posterior}.

\begin{figure*}[ht!]
    \centering
        \includegraphics[width=0.49\linewidth, bb=0 0 398 398]{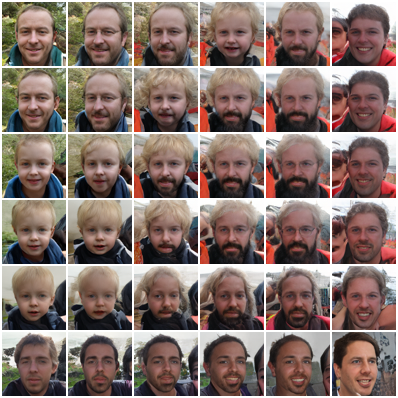}
        \includegraphics[width=0.49\linewidth, bb=0 0 398 398]{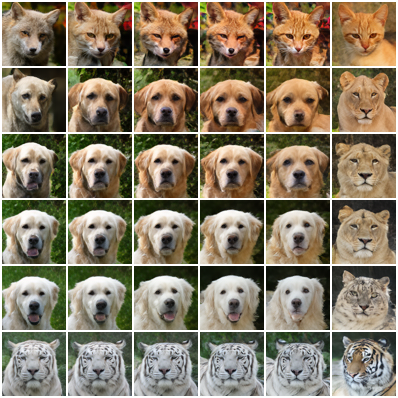}
    \caption{Latent space interpolation for $64\times 64$ generated images. The four corner images are interpolated by a weighted mixture of their latent noises, such that the other images are ``in-between'' images from the latent perspective, similar to what has been done in GANs \cite{karras2019style}.}   
    \label{fig:interp_avg}
\end{figure*}

\section{Experiments}
\label{sec:exp}

In this section, we present the evaluation of UDPM under multiple scenarios. We tested our method on CIFAR10 \cite{cifar10}, FFHQ \cite{karras2019style}, and AFHQv2 \cite{choi2020stargan} datasets. Here we focus on the qualitative performance of UDPM and demonstrate its interpolatable latent space. In the appendix we provide additional quantitative and qualitative results. 

We set $L=3$ and fix $\gamma = 2$ for all datasets. We also use a uniform box filter of size $2\times 2$ as the downsampling kernel $\w$ as it satisfies the condition in Lemma \ref{lemma:sampled_gauss}. We then normalize it w.r.t. its norm $\|\w\|$ and use it to construct the downsampling operator $\mathcal{H}$. Unlike previous diffusion approaches that are limited to analytically defined noise schedulers due to the large diffusion steps number, UDPM allows the noise scheduler to be fine-tuned manually by setting only 6 values ($\{\alpha_l\}_{l=1}^L$ and $\{\sigma_l\}_{l=1}^L$). In our tests we set $\{\alpha_l\}_{l=1}^3=\{0.5, 0.2, 10^{-3}\}$ and $\{\sigma_l\}_{l=0}^3=\{0.1, 0.2, 0.3\}$ for all datasets.

We use the same UNet architecture proposed by \cite{song2020score} and utilized in \cite{karras2022elucidating} for all diffusion steps (the specific implementation details are presented in Table \ref{table:appndx_train_configs}). We increase the number of output channels of the network by a factor of $\gamma^2$ and use the depth-to-space layer \cite{shi2016real} to rearrange the output pixels to the desired dimension, which is equivalent to a scale-up by a factor of $\gamma$. For all datasets, we train the network for 600K training steps using ADAM \cite{kingma2014adam} with learning rate and batch size set to $10^{-4}$ and 64, respectively. We save the model weights every 10K training steps and pick the model with the best Fréchet inception distance (FID) \cite{heusel2017gans}. For stabilizing the training we use exponential moving averaging (EMA) with the dampening parameter set to 0.9999. We also set $\lambda_\text{fid}=(1, 1, 0)$, $\lambda_\text{per}=(4, 4, 0)$, and $\lambda_\text{adv}=(0.2, 0.5, 1)$. Particularly, we found that setting $\lambda_\text{fid}=\lambda_\text{per}=0$ for $l=3$ achieves the best results with minimal mode-collapse. For the discriminator network, we use the discriminator architecture used in \cite{DDGAN}, which is a variant of the original discriminator network proposed by \cite{karras2019style}. We train all models using a single NVIDIA RTX A6000 GPU.

\begin{figure*}[t]
    \centering
    \begin{tikzpicture}
        \node (img1) at (0,0) {\includegraphics[width=0.45\linewidth, bb=0 0 332 200]{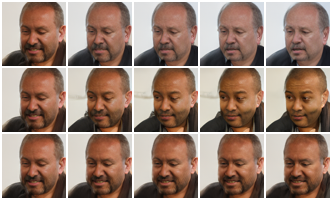}};
        \node (img2) at (7,0) {\includegraphics[width=0.45\linewidth, bb=0 0 332 200]{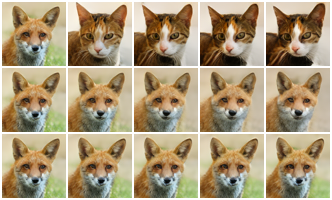}};
        \draw[thick,->] (img1.south west) -- ++(5,0) node[midway, below] {noise deviation}; 
        \draw[thick,->] (img1.south west) -- ++(0,4) node[midway, left] {$l$};  
    \end{tikzpicture}
    \caption{Latent space perturbation for $64\times 64$ generated images. The original image is on the left. To its right we present images that were generated by adding a small noise to the latent noise from diffusion step $l$. As can be seen, the initial diffusion step ($l=1$) controls the fine details of the image, while the final diffusion step ($l=3$) changes the semantics of the image.}
    \label{fig:latent_perturb}
\end{figure*}

\textbf{Unconditional Generation.}
In the unconditional scheme, we evaluate UDPM on FFHQ and AFHQv2 datasets, where each dataset contains $50K$ and $14K$ images, respectively. 
We resize the images to 64$\times$64 and train the UNet using Algorithm \ref{alg:training}.

The number of sampling steps required by UDPM is two orders of magnitude smaller than the original denoising diffusion models \cite{DDPM, sohl2015deep}, one order of magnitude less than \cite{Song2022DDIM, karras2022elucidating}, and up-to 4 times smaller than many recent sophisticated samplers \cite{lu2022dpm, lu2022dpmplus, dockhorn2022genie} designed for accelerating diffusion generation. Particularly, since the size of latent variables increases progressively when sampling, UDPM requires less total computations than a single denoising step used in traditional denoising diffusion, as we explain below (see \textbf{Runtime} paragraph). While reducing the computational costs, UDPM retains great image generation quality, as can be seen in Figures \ref{fig:FFHQ64} and \ref{fig:AFHQv264}.

In addition to the decreased number of sampling steps, UDPM has much smaller latent dimensions. This property allows us to smoothly interpolate the latent space, similarly to what has been done in GANs \cite{karras2019style}. Figures \ref{fig:interp_avg}, \ref{fig:appndx_interp_AFHQ}, and \ref{fig:appndx_interp_FFHQ} show the results, where the corner images are randomly generated and the in-between images are generated by averaging the noises used to generate the corner images: 
\begin{align*}
    \e_l(i, j) = \eta_i (\delta_j \e_1^l + \sqrt{1 - \delta_j^2} \e_2^l )
                + \sqrt{1 - \eta_i^2} (\delta_j \e_3^l + \sqrt{1 - \delta_j^2} \e_4^l),
\end{align*}
where $\e_1^l,\e_2^l,\e_3^l,\e_4^l$ are the noises at diffusion step $l$ used to generate the 4 corner images, $(i, j)$ are the indices of the interpolated image in the figure, and $(\eta_i, \delta_j)$ are the interpolation coefficients that lie in the range $[0, 1]$. In the supplementary material we investigate other interpolation regimes, such as latent variable swapping between two generated images (see Figure \ref{fig:appndx_swap}).

Another benefit of the smaller latent dimensions is that it allows better interpretability of the generative model. While in previous diffusion approaches one needed to add a conditioning mechanism to the network in order to be able to control the generation, in UDPM, one may control the generation by modifying only the noise maps of the diffusion, as can be seen in Figures \ref{fig:latent_perturb} and \ref{fig:appndx_pertrub}.

\textbf{Class Conditional Generation.} Similarly to previous works, we also propose a conditional generation scheme for UDPM, where we use the label encoding block used in \cite{karras2022elucidating} to control the generation class. We evaluate UDPM on the popular CIFAR10 \cite{cifar10} dataset, which has 50K $32\times 32$ images of 10 different classes. 

Figure \ref{fig:appndx_CIFAR10} shows the generation results of UDPM on CIFAR10. We also compare the performance of UDPM empirically to many other state-of-the-art non-distillation diffusion methods in Table \ref{table:appndx_CIFAR10_FID}, where we examine the FID results on 50K images generated from the 10 classes equally. For the baseline methods, we show the original results presented in their papers and the results when the number of steps is reduced to 5 diffusion steps. As can be seen, when the number of steps is limited to 5, UDPM outperforms all the methods, while also requiring considerably less computation; specifically, $1/3$ of a single typical diffusion step.

\begin{table}[!t]
    \centering
    \resizebox{0.4\linewidth}{!}{
    \begin{tabular}{ | c | c | c | c | c |  }
    \hline
    \centering
                                                            & steps &    FID  \\ \hline   
    DDIM \cite{Song2022DDIM}                                & 10/5  &  13.36/93.51  \\ \hline   
    DPM-Solver \cite{lu2022dpmplus}                         & 10/5  &  6.96/288.99   \\ \hline
    EDM   \cite{karras2022elucidating}                      & 35/5  &  1.79/35.54  \\ \hline
    GENIE \cite{dockhorn2022genie}                          & 5   &  11.20  \\ \hline
    DEIS  \cite{jolicoeur2021gotta}                         & 5   &  15.37   \\ \hline
    GGDM  \cite{watson2021learning}                         & 5   &  13.77  \\ \hline
    DDGAN  \cite{DDGAN}                                     & 2   &  4.08  \\ \hline \hline
    TDPM \cite{zheng2022truncated}                          & 1   &  8.91  \\ \hline
    CT \cite{song2023consistency} & 1   &  8.70   \\ \hline
    \textbf{UDPM (ours)}                                    & \textbf{<1}  &  \textbf{6.86}  \\ \hline
    
    \end{tabular}
    }
    \vspace{3mm}
\caption{FID scores on the CIFAR10 dataset 
\cite{cifar10}
. UDPM uses 3 steps, which are equivalent in terms of complexity to 0.3 of a single denoising step used in typical diffusion models like DDPM or EDM.}
\label{table:appndx_CIFAR10_FID}
\end{table}

\textbf{Runtime.}
Unlike typical denoising diffusion models, UDPM decreases the dimensions of the latent variables when proceeding with the diffusion process. Thus, the computations required to run the reverse diffusion process for sampling are significantly less extensive. Specifically, sampling a $64\times 64$ image using a single denoising diffusion step requires 40.62 GFLOPS, while UDPM samples images with the same network structure using 13.35 GFLOPs, which is equivalent to $30\%$ of the computations. The reason is that in its 3 steps, the input dimensions of the UDPM network are $8 \times 8$, $16 \times 16$ and $32 \times 32$

When comparing the runtimes, UDPM can sample $64\times 64$ images from FFHQ and AFHQv2 datasets at the rate of 765.21 FPS (frames per second) when benchmarked on NVIDIA RTX A6000 GPU, while a single denoising diffusion step can be run at the rate of 255 FPS using a very similar network.  

\textbf{Limitations.} While our approach produces impressive generative performance, it has some limitations. One limitation of our work is the evaluation of relatively small datasets, such as CIFAR10, FFHQ, and AFHQv2, which is a consequence of our limited computational resources. This restricted us from evaluating our approach on larger and more diverse datasets. Future work could overcome this limitation by leveraging a more powerful computational infrastructure to conduct experiments on larger datasets, providing a more comprehensive validation of the model's capabilities. Additionally, while UDPM offers improved interpretability over ``standard'' denoising diffusion models, it still falls short compared to the interpretability shown for GANs. To address this, further research could explore enhancing the latent space structure of UDPMs to make it more interpretable, perhaps by incorporating techniques from GANs or further analyzing the currently generated latent space.
\section{Conclusion}

We have proposed a new diffusion-based generative model called the Upsampling Diffusion Probabilistic Model (UPDM). Our approach reduces the number of diffusion steps required to produce high-quality images, which makes it significantly more efficient than previous solutions.

We have demonstrated the effectiveness of our approach on three different datasets: CIFAR10, FFHQ, and AFHQv2. It is capable of producing high-quality images with only 3 diffusion steps, whose computational cost is less than for one step of the original diffusion models, while significantly outperforming current state-of-the-art methods dedicated to diffusion sampling acceleration.

Furthermore, we have shown that our interpolatable latent space has potential for further exploration, particularly in the realm of image editing. Based on the initial results shown in the paper, it contains semantic directions, such as making people smile or changing their age as shown in Figure~\ref{fig:latent_perturb}. Future research may explore using UDPM to perform editing operations similar to those performed in styleGAN while maintaining better generation capabilities and favorable diffusion properties.

\clearpage


\bibliographystyle{plain}
\bibliography{egbib}

\newpage
\newpage
\appendix
\onecolumn

\section*{UDPM: Upsampling Diffusion Probabilistic Models -- Supplementary Material}

Here we provide our social impage statement, extended derivations of our UDPM model, ablation studies on the latent space, and additional generation results.

\section{Social Impact Statement}

The development of Upsampling Diffusion Probabilistic Models (UDPM) represents a significant advancement in the field of image generation, offering the ability to produce high-quality images with fewer computational resources. However, as with any powerful technology, there are potential risks and ethical considerations to address. One major concern is the potential misuse of this technology for creating realistic but deceptive images, such as deepfakes, which can be used to spread misinformation or for malicious purposes. Additionally, the ability to generate high-quality synthetic images raises issues of copyright and ownership, potentially impacting artists and content creators whose work might be replicated or modified without their consent.

To mitigate these risks, it is essential to implement safeguards and establish ethical guidelines for the use of UDPM and similar technologies. This can include developing robust detection methods for synthetic content, ensuring transparency in the creation and distribution of AI-generated media, and promoting responsible use among developers and end-users. Collaborating with policymakers to create regulations that address the misuse of generative models can also help in preventing harm. Furthermore, fostering an open dialogue within the AI research community about the ethical implications and potential societal impacts of such technologies will be crucial in ensuring that advancements in this field are aligned with broader societal values and public interest.

\section{Extended Derivations}
\label{sec:ext_derivations}

\subsection{Evidence Lower Bound}\label{appndx:ELBO}

From the Evidence Lower Bound (ELBO) we have $\mathbb{E}[-\log p_\theta(\x_0)] \leq \mathbb{E}_q\left[ \log\frac{p_\theta (\x_{0:L})}{q(\x_{1:L}|\x_0)}\right]$, hence
\begin{align*}
    \mathbb{E}[-\log p_\theta(\x_0)] &\leq \mathbb{E}_q\left[ \log\frac{p_\theta (\x_{0:L})}{q(\x_{1:L}|\x_0)}\right] = \mathbb{E}_q \left[ -\log p(\x_L) - \sum_{l=1}^L\log\frac{p_\theta(\x_{l-1}|\x_l)}{q(\x_l|\x_{l-1})}\right] \\ 
    &= \mathbb{E}_q \left[ -\log p(\x_L) - \sum_{l=2}^L\log \frac{p_\theta(\x_{l-1}|\x_l)}{q(\x_l|\x_{l-1})} - \log \frac{p_\theta(\x_0|\x_1)}{q(\x_1|\x_0)}\right] \\
    &\stackrel{(\star)}{=} \mathbb{E}_q \left[ -\log p(\x_L) - \sum_{l=2}^L\log \frac{p_\theta(\x_{l-1}|\x_l) q(\x_{l-1}|\x_l)}{q(\x_{l-1}|\x_{l}, \x_0)q(\x_{l}|\x_0)} - \log \frac{p_\theta(\x_0|\x_1)}{q(\x_1|\x_0)}\right] \\ 
    &= \mathbb{E}_q \left[ -\log p(\x_L) - \sum_{l=2}^L\log \frac{p_\theta(\x_{l-1}|\x_l)}{q(\x_{l-1}|\x_{l}, \x_0)} - \sum_{l=2}^L \log q(\x_{l-1}|\x_0) + \sum_{l=2}^L \log q(\x_{l}|\x_0) - \log \frac{p_\theta(\x_0|\x_1)}{q(\x_1|\x_0)}\right] \\
    &= \mathbb{E}_q \left[ -\log p(\x_L) - \sum_{l=2}^L\log \frac{p_\theta(\x_{l-1}|\x_l)}{q(\x_{l-1}|\x_{l}, \x_0)} - \cancel{\log q(\x_{1}|\x_0)} + \log q(\x_{L}|\x_0) - \log \frac{p_\theta(\x_0|\x_1)}{\cancel{q(\x_1|\x_0)}}\right] \\
    &= \mathbb{E}_q \left[ -\log \frac{p(\x_L)}{q(\x_{L}|\x_0)} - \sum_{l=2}^L\log \frac{p_\theta(\x_{l-1}|\x_l)}{q(\x_{l-1}|\x_{l}, \x_0)} - \log p_\theta(\x_0|\x_1)\right] \\
    &= \mathbb{E}_q \left[ D_\text{KL}(p(\x_L)||q(\x_{L}|\x_0)) + \sum_{l=2}^LD_\text{KL}(p_\theta(\x_{l-1}|\x_l)||q(\x_{l-1}|\x_{l}, \x_0)) - \log p_\theta(\x_0|\x_1)\right],
\end{align*}
where in $(\star)$ Bayes was used, particularly 
\begin{equation*}
    q(\x_l|\x_{l-1})=q(\x_l|\x_{l-1}, \x_0) = \frac{q(\x_{l-1}, \x_l| \x_0)}{q(\x_{l-1}|\x_0)} = \frac{q(\x_{l-1}|\x_l, \x_0) q(\x_l|\x_0)}{q(\x_{l-1}|\x_0)}.
\end{equation*}

\subsection{Lemma \ref{lemma:sampled_gauss}} \label{appndx:lemma_proof}
\begin{proof}
The characteristic function of a Normal vector $\e \sim \mathcal{N}(\mu, \Sigma)$ is in the form
\begin{equation}
    \phi_\e(\mathbf{t}) = \mathbb{E}[\exp(i \mathbf{t}^T \e)] = \exp (i\mathbf{t}^T \mu - \frac{1}{2}\mathbf{t}^T\Sigma \mathbf{t}),\label{eq:Gauss_char}
\end{equation}
when $\mu=0$ and $\Sigma = \I$ we get
\begin{equation*}
    \phi_\e(\mathbf{t}) = \exp ( - \frac{1}{2}\mathbf{t}^T \mathbf{t}).
\end{equation*}

To prove the first claim of the lemma, it is sufficient to prove that $\mathcal{H}\e$ has a characteristic function of the form \eqref{eq:Gauss_char}. Denote the transpose operator of $\mathcal{H}$ by $\mathcal{H}^T$, which is defined as zero padding operator followed by applying a flipped version the kernel of $\mathcal{H}$, denoted by $\w$. We have
\begin{align*}
    \phi_{\mathcal{H}\e}(\mathbf{t}) &= \mathbb{E}[\exp (i \mathbf{t}^T(\mathcal{H} \e))] = \exp (i (\mathcal{H}^T \mathbf{t})^T\e) = \exp (i(\mathcal{H}^T \mathbf{t})^T \mu - \frac{1}{2}(\mathcal{H}^T \mathbf{t})^T\Sigma (\mathcal{H}^T \mathbf{t})) \\
    &= \exp (i\mathbf{t}^T\mathcal{H}  \mu - \frac{1}{2}\mathbf{t}^T\mathcal{H} \Sigma \mathcal{H}^T \mathbf{t})\stackrel{\{\mu=0, \Sigma=I\}}{=} \exp (- \frac{1}{2}\mathbf{t}^T\mathcal{H}\mathcal{H}^T \mathbf{t}),\\
    &\Rightarrow \mathcal{H}\e \sim \mathcal{N}(0, \mathcal{H}\mathcal{H}^T).
\end{align*}

All that remains is to express $\mathcal{H}\mathcal{H}^T$ under the assumption on the support of the downsampling kernel . The operator $\mathcal{H}$ is defined as applying blur kernel $\w = [w_{-\lfloor\gamma/2\rfloor + 1}, \hdots, w_0, \hdots w_{\lfloor\gamma/2\rfloor}]$ followed by subsampling with stride $\gamma$, which can be represented in matrix form by $\mathcal{H} = \mathcal{S}_\gamma \mathcal{W}$, where $\mathcal{S}_\gamma\in\mathbb{R}^{M\times N}$ and $\mathcal{W}\in\mathbb{R}^{N\times N}$. Specifically 
\begin{equation*}
    \mathcal{S}_\gamma = 
    \begin{pmatrix}
        1\ 0\ \hdots\ 0\ 0\ \hdots\ 0& \hdots & 0 \\
        \underbrace{0\ \hdots\ 0}_\gamma\ 1\ 0\ \hdots\ 0 & \hdots  & 0\\
        \underbrace{0\ \hdots\ 0}_\gamma\ \underbrace{0\ \hdots\ 0}_\gamma\  1& \hdots  & 0\\
        \vdots & \vdots & \vdots
    \end{pmatrix},  \\
\end{equation*}

\begin{equation*}
    \mathcal{W}_\gamma = 
    \begin{pmatrix}
        w_0 & w_1 & \hdots & w_{\lfloor\gamma/2\rfloor} & 0 &\hdots  & w_{-\lfloor\gamma/2\rfloor + 1} & w_{-\lfloor\gamma/2\rfloor + 2} &   \hdots & w_{-1} \\
        w_{-1} & w_0 & \hdots & w_{\lfloor\gamma/2\rfloor} & 0 &\hdots  & 0 & w_{-\lfloor\gamma/2\rfloor + 1} & \hdots & w_{-2} \\
        \vdots & & & \vdots & & &  \vdots & & & \vdots
    \end{pmatrix},
\end{equation*}
Therefore we have
\begin{align}
    \mathcal{S}_\gamma \mathcal{W} = 
    \begin{pmatrix}
        w_0 &  \hdots & w_{\lfloor\gamma/2\rfloor} & 0 &\hdots &0 &\hdots  & 0 &w_{-\lfloor\gamma/2\rfloor + 1}  &   \hdots \\
        0 & \hdots & 0 & w_{-\lfloor\gamma/2\rfloor + 1} & \hdots  & w_{\lfloor\gamma/2\rfloor} & \hdots & 0  & 0&\hdots    \\
        \vdots & & & \vdots & & &  \vdots & & & \vdots
    \end{pmatrix},
    \label{appndx:lemma_proof_SW}
\end{align}
as can be observed from \eqref{appndx:lemma_proof_SW}, the rows of $\mathcal{S_\gamma\mathcal{W}}$ do not intersect with each other, as a result we get
\begin{align}
    \mathcal{H} \mathcal{H}^T = \mathcal{S}_\gamma \mathcal{W} (\mathcal{S}_\gamma \mathcal{W})^T = \mathcal{S}_\gamma \mathcal{W} \mathcal{W}^T \mathcal{S}_\gamma^T =
    \begin{pmatrix}
        \|\w\|_2^2 & 0 & 0 & \hdots & 0 \\
        0 & \|\w\|_2^2 & 0 & \hdots & 0 \\
        \vdots & \vdots & \vdots & \hdots & \vdots \\
        0 & 0 & 0 & \hdots & \|\w\|_2^2
    \end{pmatrix} 
    = \|\w\|_2^2 \I.
\end{align}

\end{proof}

\subsection{Efficient representation of the forward process}\label{appndx:q_xt_given_x0}
Similar to earlier works, for efficient training, one would like to sample $\x_l$ directly using $\x_0$. Which is viable when assuming that the support of $\w$ is at most $\gamma$ and $\|w\|_2^2 = 1$. Formally
\begin{align*}
    \x_l &= \alpha_l \mathcal{H} \x_{l-1} + \sigma_l \e_1 = \alpha_l \mathcal{H} (\alpha_{l-1} \mathcal{H} \x_{l-2} + \sigma_{l-1} \e_2) + \sigma_l \e_1 \\
    &= \alpha_l \alpha_{l-1} \mathcal{H}^2 \x_{l-2} + \alpha_l \sigma_{l-1} \mathcal{H} \e_2 + \sigma_l\e_1 \\
    &\stackrel{\text{Lemma } \eqref{lemma:sampled_gauss}}{=}
    \alpha_l \alpha_{l-1} \mathcal{H}^2 \x_{l-2} + \alpha_l \sigma_{l-1} \tilde{\e}_2 + \sigma_l\e_1,
\end{align*}
where $\e_1, \e_2, \tilde{\e}_2 \stackrel{i.i.d.}{\sim} \mathcal{N}(0, \I)$. By definition, $\e_1$ and $\tilde{\e}_2$ are independent identically distributed vectors, therefore one may write
\begin{align*}
    \x_l   &= \alpha_l \alpha_{l-1} \mathcal{H}^2 \x_{l-2} + \sqrt{\alpha_l^2 \sigma_{l-1}^2 + \sigma_l^2}\e,
\end{align*}
repeating the opening $l-2$ times leads to
\begin{align*}
    \x_l &= \bar{\alpha}_{l}\mathcal{H}^l \x_{0} + \sqrt{\alpha_l^2\alpha_{l-1}^2...\alpha_2^2\sigma^2_1+\alpha_l^2\alpha_{l-1}^2...\alpha_3^2\sigma^2_2+...+\alpha_l^2\sigma_{l-1}^2 + \sigma_l^2}\e \\
    &= \bar{\alpha}_{l}\mathcal{H}^l \x_{0} + \sqrt{\frac{\bar{\alpha}_{l}^2\sigma^2_1}{\bar{\alpha}_{1}}+\frac{\bar{\alpha}_{l}^2\sigma^2_2}{\bar{\alpha}_{2}}+...+\frac{\bar{\alpha}_{l}^2\sigma^2_1}{\bar{\alpha}_{l}}}\e = \bar{\alpha}_{l}\mathcal{H}^l \x_{0} + \bar{\alpha}_{l}\sqrt{\frac{\sigma^2_1}{\bar{\alpha}_{1}}+\frac{\sigma^2_2}{\bar{\alpha}_{2}}+...+\frac{\sigma^2_1}{\bar{\alpha}_{l}}}\e \\
    &= \bar{\alpha}_{l}\mathcal{H}^l \x_{0} + \tilde{\sigma}_l\e,
\end{align*}
where $\bar{\alpha}_l = \prod_{k=1}^L \alpha_k$, $\tilde{\sigma}_l^2=\bar{\alpha}_l^2\sum_{k=1}^l\frac{\sigma_k^2}{\bar{\alpha}_k^2}$, and $\e \sim \mathcal{N}(0, \I)$.

\subsection{Derivation of the posterior}\label{appndx:q_xt1_given_xt_x0}
Using Bayes theorem, we have for $l > 1$
\begin{align*}
    q(\x_{l-1}|\x_l, \x_0) &= \frac{q(\x_{l-1}, \x_l| \x_0)}{q(\x_l|\x_0)} = \frac{q(\x_l|\x_{l-1},\cancel{\x}_0) q(\x_{l-1}|\x_0)}{q(\x_l|\x_0)} \stackrel{\text{Markov}}{=} \frac{q(\x_l|\x_{l-1}) q(\x_{l-1}|\x_0)}{q(\x_l|\x_0)} \\
    &\stackrel{(\star\star)}{\propto} \exp \left( -\frac{1}{2 \sigma_l^2}\|\x_l - \alpha_l\mathcal{H}\x_{l-1}\|_2^2 -\frac{1}{2\tilde{\sigma}_{l-1}^2} \|\x_{l-1} - \bar{\alpha}_{l-1}\mathcal{H}^{l-1}\x_0\|_2^2\right)  \\
    &= \exp( -\frac{1}{2 \sigma_l^2}(\x_l^T\x_l - 2\alpha_l\x_{l-1}^T \mathcal{H}^T \x_l + \alpha_l^2 \x_{l-1}^T \mathcal{H}^T\mathcal{H} \x_{l-1}) \\
    &-\frac{1}{2\tilde{\sigma}_{l-1}^2} (\x_{l-1}^T\x_{l-1} 
     -2\bar{\alpha}_{l-1} \x_{l-1}^T\mathcal{H}^{l-1} \x_0 + \bar{\alpha}_{l-1}^2 \x_0^T(\mathcal{H}^{l-1})^T \mathcal{H}^{l-1}\x_0))  \\
    &\stackrel{(\star\star)}{\propto} \exp\left\{ -\frac{1}{2}\x_{l-1}^T\left(\frac{\alpha_l^2}{\sigma_l^2}\mathcal{H}^T\mathcal{H} + \frac{1}{\tilde{\sigma}_{l-1}^2}\I\right)\x_{l-1} + \x_{l-1}^T\left(\frac{\alpha_l}{\tilde{\sigma}_l^2}\mathcal{H}^T\x_{l} + \frac{\bar{\alpha}_{l-1}}{\tilde{\sigma}_{l-1}^2}\mathcal{H}^{l-1}\x_0\right) \right\},
\end{align*}
which is a Normal distribution form w.r.t. the random vector $\x_{l-1}$. Note that in $(\star\star)$ the proportional equivalence is with relation to $\x_{l-1}$.
Let $q(\x_{l-1}|\x_l, \x_0)=\mathcal{N}(\mu, \Sigma)$, then one may write
\begin{equation}
    \Sigma^{-1} = \frac{\alpha_l^2}{\sigma_l^2}\mathcal{H}^T\mathcal{H} + \frac{1}{\tilde{\sigma}_{l-1}^2} \I, \label{eq:sigma_1_appndx}
\end{equation}
and
\begin{equation}
    \Sigma^{-1} \mu = \frac{\alpha_{l}}{\sigma_l^2}\mathcal{H}^T\x_l + \frac{\bar{\alpha}_{l-1}}{\tilde{\sigma}_{l-1}^2} \mathcal{H}^{l-1}\x_0. \label{eq:Sigma_1_mu_appndx}
\end{equation}

\subsection{Mean and variance of the posterior}\label{appndx:Sigma_mu_derivation}
The terms in \eqref{eq:sigma_1_appndx} and \eqref{eq:Sigma_1_mu_appndx} seem hard to evaluate, however, in the following we present an efficient way to compute them. By definition $\mathcal{H}$ is structured from a circular convolution with a blur filter followed by a subsampling with stride $\alpha$. Therefore, one may use the poly-phase identity used in \cite{corrFlt, chan2016plug}, which states that $\mathcal{H}^T\mathcal{H}$ in this case is a circular convolution between the blur kernel $\w$ and its flipped version $\tilde{\w}$ followed by subsampling with factor $\alpha$, formally
\begin{equation*}
    \h = (\w \circledast flip(\w))\downarrow_\alpha,
\end{equation*}
therefore, one may write $\mathcal{H}^T\mathcal{H}$  in the following form
\begin{equation}
    \mathcal{H}^T\mathcal{H} = \mathcal{F}^\star \Lambda_\h \mathcal{F}, \label{eq:HTH_appndx}
\end{equation}
where $\mathcal{F}$ is the Discrete Fourier Transform (DFT), $\mathcal{F}^\star$ is the inverse DFT, and $\Lambda_\h$ is a diagonal operator representing the DFT transform of $\h$. By plugging \eqref{eq:HTH_appndx} into \eqref{eq:sigma_1_appndx} we get
\begin{align*}
    \Sigma^{-1} = \frac{\alpha_l^2}{\sigma_l^2}\mathcal{F}^\star \Lambda_\h \mathcal{F} + \frac{1}{\tilde{\sigma}_{l-1}^2} \I = \frac{\alpha_l^2}{\sigma_l^2}\mathcal{F}^\star \Lambda_\h \mathcal{F} + \frac{1}{\tilde{\sigma}_{l-1}^2} \mathcal{F}^\star\mathcal{F} =  \mathcal{F}^\star \left(\frac{\alpha_l^2}{\sigma_l^2} \Lambda_\h + \frac{1}{\tilde{\sigma}_{l-1}^2} \I \right) \mathcal{F},
\end{align*}
equivalently 
\begin{align}
    \Sigma = \mathcal{F}^\star \left(\frac{\alpha_l^2}{\sigma_l^2} \Lambda_\h + \frac{1}{\tilde{\sigma}_{l-1}^2} \I \right)^{-1} \mathcal{F}. \label{eq:sigma_appndx}
\end{align}

Therefore, as can be observed from \eqref{eq:sigma_appndx}, applying $\Sigma$ is equivalent to applying the inverse of the filter $\left(\frac{\alpha_l^2}{\sigma_l^2} \Lambda_\h + \frac{1}{\tilde{\sigma}_{l-1}^2} \I \right)$, which can be performed efficiently using Fast Fourier Transform (FFT). Finally we have
\begin{equation}
    \mu = \Sigma \left( \frac{\alpha_{l}}{\sigma_l^2}\mathcal{H}^T\x_l + \frac{\bar{\alpha}_{l-1}}{\tilde{\sigma}_{l-1}^2} \mathcal{H}^{l-1}\x_0 \right)
\end{equation}

\subsection{Sampling the posterior}\label{appndx:sampling_posterior}
As discussed in section \ref{sec:generation_using_UDPM}, the proposed generation scheme requires to sample from $\mathcal{N}(\mu_l, \Sigma_l)$ at each diffusion step, which due to the dimensions of $\Sigma_l$ is not naive, since $\Sigma_l^{0.5}$ is needed in order to use the parameterization $\x_{l-1}|\x_l = \mu_l(\x_l) + \Sigma_l^{0.5} \e$ where $\e \sim \mathcal{N}(0, I)$. However, as we saw in section \ref{appndx:Sigma_mu_derivation}, the operator $\Sigma_l$ is equivalent to applying a linear filter $\h$, therefore one may seek to find a filter $\mathbf{d}$ such that
\begin{equation*}
    \Sigma = \mathcal{F}^\star \Lambda_\h \mathcal{F} = \mathcal{F}^\star \Lambda_\mathbf{d}^2 \mathcal{F}
\end{equation*}
equivalently, we can break the DFT of the filter $\mathbf{h}$ to magnitude and phase such that
\begin{equation*}
    \text{DFT}\{\h\} = |\text{DFT}\{\h\}| e^{i\phi_h} = |\text{DFT}\{\mathbf{d}\}|^2 e^{i2\phi_d}
\end{equation*}
therefore
\begin{equation}
    |\text{DFT}\{\mathbf{d}\}| = \sqrt{|\text{DFT}\{\mathbf{h}\}|},\ \measuredangle |\text{DFT}\{\mathbf{d}\}| = \frac{1}{2} \arctan \frac{Im\{\text{DFT}\{\mathbf{h}\}\}}{Re\{\text{DFT}\{\mathbf{h}\}\}} \label{eq:d_appndx}
\end{equation}
As a result, applying $\Sigma^{0.5}$ can be performed by employing the filter $\mathbf{d}$ defined in \eqref{eq:d_appndx}.

\section{Additional results}
\label{sec:appndx_add_results} 

\subsection{Latent space interpolation}
\begin{figure*}[!bh]
    \centering
        \includegraphics[width=\linewidth, bb=0 0 398 398, trim={0 0 0 0}, clip]{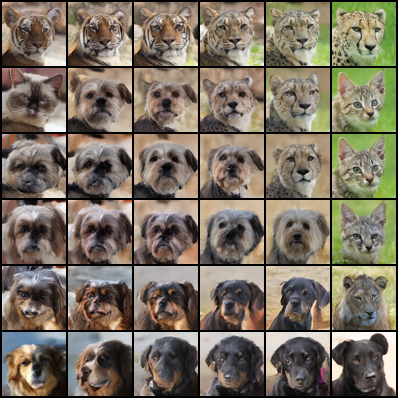}
    \caption{AFHQv2 \cite{choi2020stargan} latent space interpolation example. The four corner images are interpolated by a weighted mixture of their latent noises, such that the other images are ``in-between'' images from the latent perspective, similar to what has been done in GANs \cite{karras2019style}. All the images are of size $64\times 64$.}   
    \label{fig:appndx_interp_AFHQ}
\end{figure*}

\begin{figure*}[!h]
    \centering
        \includegraphics[width=\linewidth, bb=0 0 398 398, trim={0 0 0 0}, clip]{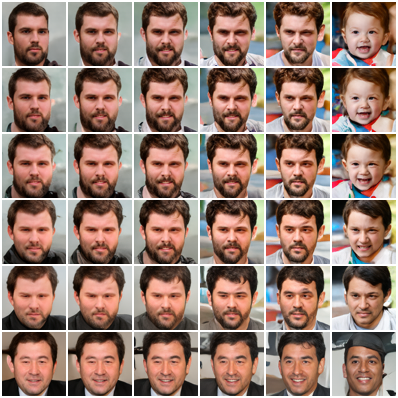}
    \caption{FFHQ \cite{karras2019style} latent space interpolation example. The four corner images are interpolated by a weighted mixture of their latent noises, such that the other images are ``in-between'' images from the latent perspective, similar to what has been done in GANs \cite{karras2019style}. All the images are of size $64\times64$.}   
    \label{fig:appndx_interp_FFHQ}
\end{figure*}

As shown previously by \cite{karras2019style}, generative models have a very interesting property, where one may modify the latent variable of the generative model in order to control the ``style'' of the generated result, or for instance, given the latent variables of two images, one may interpolate the latent space and get an ``in-between'' images that in contrast to the pixel-domain interpolation, transition smoothly and naturally, as can be seen in Figures \ref{fig:appndx_interp_AFHQ} and \ref{fig:appndx_interp_FFHQ}.

\subsection{Latent space perturbation}
\begin{figure*}[!ht]
    \centering
    \begin{tikzpicture}
        \node (img1) at (0,0) {\includegraphics[width=0.45\linewidth, bb=0 0 332 200]{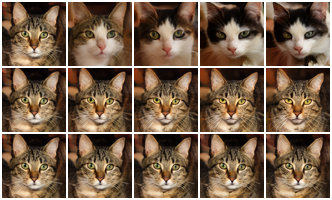}};
        \node (img2) at (7,0) {\includegraphics[width=0.45\linewidth, bb=0 0 332 200]{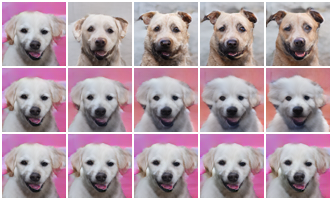}};
        \node (img3) at (0,4.5) {\includegraphics[width=0.45\linewidth, bb=0 0 332 200]{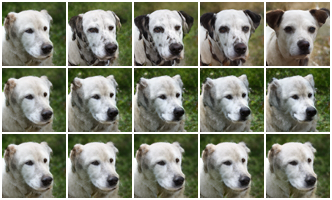}};
        \node (img4) at (7,4.5) {\includegraphics[width=0.45\linewidth, bb=0 0 332 200]{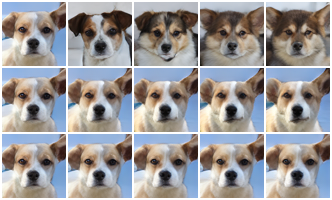}};
        \node (img5) at (0,9) {\includegraphics[width=0.45\linewidth, bb=0 0 332 200]{FFHQ_perturb/perturb_13_138.png}};
        \node (img6) at (7,9) {\includegraphics[width=0.45\linewidth, bb=0 0 332 200]{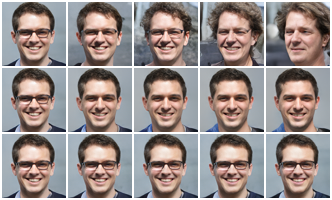}};
        \node (img5) at (0,13.5) {\includegraphics[width=0.45\linewidth, bb=0 0 332 200]{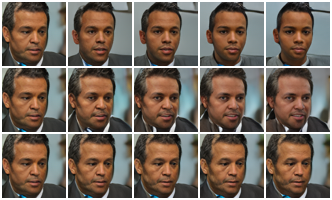}};
        \node (img6) at (7,13.5) {\includegraphics[width=0.45\linewidth, bb=0 0 332 200]{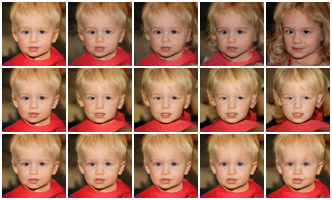}};
        \draw[thick,->] (img1.south west) -- ++(10,0) node[midway, below] {noise deviation}; 
        \draw[thick,->] (img1.south west) -- ++(0,7) node[midway, left] {$l$};  
    \end{tikzpicture}
    \caption{Latent space perturbation for $64\times 64$ generated images. The original image is on the left, then the images generated by adding a small noise to the latent noise from diffusion step $l$. As can be seen, the initial diffusion steps ($l=1$) controls the fine details of the image, while the final diffusion step ($l=3$) changes the semantics of the image.}
    \label{fig:appndx_pertrub}
\end{figure*}

Similar to what is presented in the main paper, we show additional latent variable perturbation in order to obtain better understanding of the latent space, where we add a small noise to a single diffusion step and see how it affects the generated result, as can be seen in Figure \ref{fig:appndx_pertrub}.

\subsection{Latent variable swapping}
\begin{figure*}[!ht]
    \centering
        $l=3$ \hspace{0.09\linewidth} $l=2$ \hspace{0.09\linewidth} $l=1$
        \includegraphics[width=0.8\linewidth, bb=0 0 332 68, trim={0 0 0 0}, clip]{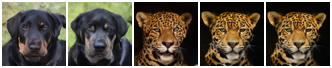}\\
        \includegraphics[width=0.8\linewidth, bb=0 0 332 68, trim={0 0 0 0}, clip]{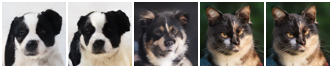}\\
        \includegraphics[width=0.8\linewidth, bb=0 0 332 68, trim={0 0 0 0}, clip]{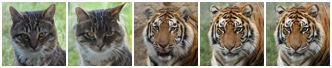}\\
        \includegraphics[width=0.8\linewidth, bb=0 0 332 68, trim={0 0 0 0}, clip]{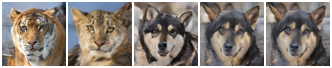} \\
        \includegraphics[width=0.8\linewidth, bb=0 0 332 68, trim={0 0 0 0}, clip]{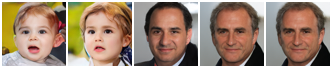}\\
        \includegraphics[width=0.8\linewidth, bb=0 0 332 68, trim={0 0 0 0}, clip]{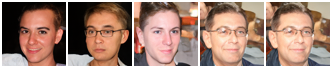}\\
        \includegraphics[width=0.8\linewidth, bb=0 0 332 68, trim={0 0 0 0}, clip]{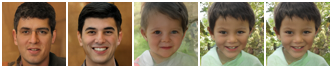}\\
        \includegraphics[width=0.8\linewidth, bb=0 0 332 68, trim={0 0 0 0}, clip]{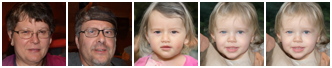}\\
        $l=3$ \hspace{0.09\linewidth} $l=2$ \hspace{0.09\linewidth} $l=1$
    \caption{Latent variable swapping: Given the left and right images with the noise maps used for generating them, we replace the $l$-th noise map of the image on the right with the $l$-th noise map of the image on the left to see how each diffusion step affect the result (middle columns).}   
    \label{fig:appndx_swap}
\end{figure*}

In addition to what we presented above, we provide an additional ablation study that shows the benefit of UDPM, that is to swap the latent variables (noise maps) between two generated images, and study the effect on the generated result, as can be seen in Figure \ref{fig:appndx_swap}.

\subsection{Generation Results}

\begin{figure*}[!ht]
    \centering      
\includegraphics[width=0.055\linewidth, bb=0 0 32 32]{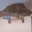}
\includegraphics[width=0.055\linewidth, bb=0 0 32 32]{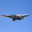}
\includegraphics[width=0.055\linewidth, bb=0 0 32 32]{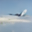}
\includegraphics[width=0.055\linewidth, bb=0 0 32 32]{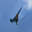}
\includegraphics[width=0.055\linewidth, bb=0 0 32 32]{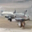}
\includegraphics[width=0.055\linewidth, bb=0 0 32 32]{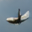}
\includegraphics[width=0.055\linewidth, bb=0 0 32 32]{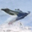}
\includegraphics[width=0.055\linewidth, bb=0 0 32 32]{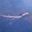}
\includegraphics[width=0.055\linewidth, bb=0 0 32 32]{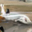}
\includegraphics[width=0.055\linewidth, bb=0 0 32 32]{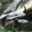}
\includegraphics[width=0.055\linewidth, bb=0 0 32 32]{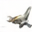}
\includegraphics[width=0.055\linewidth, bb=0 0 32 32]{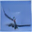}
\includegraphics[width=0.055\linewidth, bb=0 0 32 32]{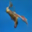}
\includegraphics[width=0.055\linewidth, bb=0 0 32 32]{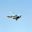}
\includegraphics[width=0.055\linewidth, bb=0 0 32 32]{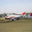}
\includegraphics[width=0.055\linewidth, bb=0 0 32 32]{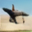}\\
\includegraphics[width=0.055\linewidth, bb=0 0 32 32]{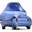}
\includegraphics[width=0.055\linewidth, bb=0 0 32 32]{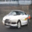}
\includegraphics[width=0.055\linewidth, bb=0 0 32 32]{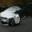}
\includegraphics[width=0.055\linewidth, bb=0 0 32 32]{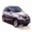}
\includegraphics[width=0.055\linewidth, bb=0 0 32 32]{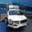}
\includegraphics[width=0.055\linewidth, bb=0 0 32 32]{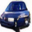}
\includegraphics[width=0.055\linewidth, bb=0 0 32 32]{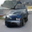}
\includegraphics[width=0.055\linewidth, bb=0 0 32 32]{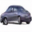}
\includegraphics[width=0.055\linewidth, bb=0 0 32 32]{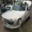}
\includegraphics[width=0.055\linewidth, bb=0 0 32 32]{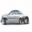}
\includegraphics[width=0.055\linewidth, bb=0 0 32 32]{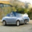}
\includegraphics[width=0.055\linewidth, bb=0 0 32 32]{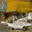}
\includegraphics[width=0.055\linewidth, bb=0 0 32 32]{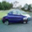}
\includegraphics[width=0.055\linewidth, bb=0 0 32 32]{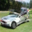}
\includegraphics[width=0.055\linewidth, bb=0 0 32 32]{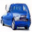}
\includegraphics[width=0.055\linewidth, bb=0 0 32 32]{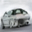}\\
\includegraphics[width=0.055\linewidth, bb=0 0 32 32]{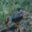}
\includegraphics[width=0.055\linewidth, bb=0 0 32 32]{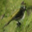}
\includegraphics[width=0.055\linewidth, bb=0 0 32 32]{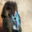}
\includegraphics[width=0.055\linewidth, bb=0 0 32 32]{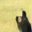}
\includegraphics[width=0.055\linewidth, bb=0 0 32 32]{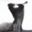}
\includegraphics[width=0.055\linewidth, bb=0 0 32 32]{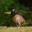}
\includegraphics[width=0.055\linewidth, bb=0 0 32 32]{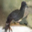}
\includegraphics[width=0.055\linewidth, bb=0 0 32 32]{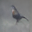}
\includegraphics[width=0.055\linewidth, bb=0 0 32 32]{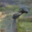}
\includegraphics[width=0.055\linewidth, bb=0 0 32 32]{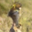}
\includegraphics[width=0.055\linewidth, bb=0 0 32 32]{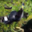}
\includegraphics[width=0.055\linewidth, bb=0 0 32 32]{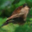}
\includegraphics[width=0.055\linewidth, bb=0 0 32 32]{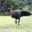}
\includegraphics[width=0.055\linewidth, bb=0 0 32 32]{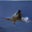}
\includegraphics[width=0.055\linewidth, bb=0 0 32 32]{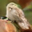}
\includegraphics[width=0.055\linewidth, bb=0 0 32 32]{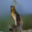}\\
\includegraphics[width=0.055\linewidth, bb=0 0 32 32]{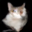}
\includegraphics[width=0.055\linewidth, bb=0 0 32 32]{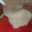}
\includegraphics[width=0.055\linewidth, bb=0 0 32 32]{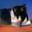}
\includegraphics[width=0.055\linewidth, bb=0 0 32 32]{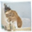}
\includegraphics[width=0.055\linewidth, bb=0 0 32 32]{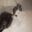}
\includegraphics[width=0.055\linewidth, bb=0 0 32 32]{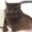}
\includegraphics[width=0.055\linewidth, bb=0 0 32 32]{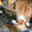}
\includegraphics[width=0.055\linewidth, bb=0 0 32 32]{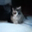}
\includegraphics[width=0.055\linewidth, bb=0 0 32 32]{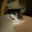}
\includegraphics[width=0.055\linewidth, bb=0 0 32 32]{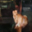}
\includegraphics[width=0.055\linewidth, bb=0 0 32 32]{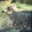}
\includegraphics[width=0.055\linewidth, bb=0 0 32 32]{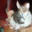}
\includegraphics[width=0.055\linewidth, bb=0 0 32 32]{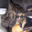}
\includegraphics[width=0.055\linewidth, bb=0 0 32 32]{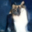}
\includegraphics[width=0.055\linewidth, bb=0 0 32 32]{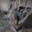}
\includegraphics[width=0.055\linewidth, bb=0 0 32 32]{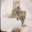}\\
\includegraphics[width=0.055\linewidth, bb=0 0 32 32]{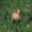}
\includegraphics[width=0.055\linewidth, bb=0 0 32 32]{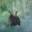}
\includegraphics[width=0.055\linewidth, bb=0 0 32 32]{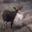}
\includegraphics[width=0.055\linewidth, bb=0 0 32 32]{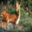}
\includegraphics[width=0.055\linewidth, bb=0 0 32 32]{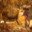}
\includegraphics[width=0.055\linewidth, bb=0 0 32 32]{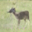}
\includegraphics[width=0.055\linewidth, bb=0 0 32 32]{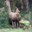}
\includegraphics[width=0.055\linewidth, bb=0 0 32 32]{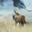}
\includegraphics[width=0.055\linewidth, bb=0 0 32 32]{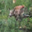}
\includegraphics[width=0.055\linewidth, bb=0 0 32 32]{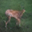}
\includegraphics[width=0.055\linewidth, bb=0 0 32 32]{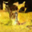}
\includegraphics[width=0.055\linewidth, bb=0 0 32 32]{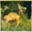}
\includegraphics[width=0.055\linewidth, bb=0 0 32 32]{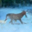}
\includegraphics[width=0.055\linewidth, bb=0 0 32 32]{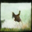}
\includegraphics[width=0.055\linewidth, bb=0 0 32 32]{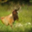}
\includegraphics[width=0.055\linewidth, bb=0 0 32 32]{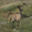}\\
\includegraphics[width=0.055\linewidth, bb=0 0 32 32]{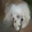}
\includegraphics[width=0.055\linewidth, bb=0 0 32 32]{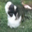}
\includegraphics[width=0.055\linewidth, bb=0 0 32 32]{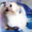}
\includegraphics[width=0.055\linewidth, bb=0 0 32 32]{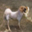}
\includegraphics[width=0.055\linewidth, bb=0 0 32 32]{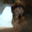}
\includegraphics[width=0.055\linewidth, bb=0 0 32 32]{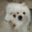}
\includegraphics[width=0.055\linewidth, bb=0 0 32 32]{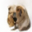}
\includegraphics[width=0.055\linewidth, bb=0 0 32 32]{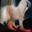}
\includegraphics[width=0.055\linewidth, bb=0 0 32 32]{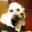}
\includegraphics[width=0.055\linewidth, bb=0 0 32 32]{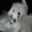}
\includegraphics[width=0.055\linewidth, bb=0 0 32 32]{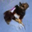}
\includegraphics[width=0.055\linewidth, bb=0 0 32 32]{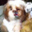}
\includegraphics[width=0.055\linewidth, bb=0 0 32 32]{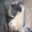}
\includegraphics[width=0.055\linewidth, bb=0 0 32 32]{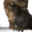}
\includegraphics[width=0.055\linewidth, bb=0 0 32 32]{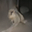}
\includegraphics[width=0.055\linewidth, bb=0 0 32 32]{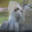}\\
\includegraphics[width=0.055\linewidth, bb=0 0 32 32]{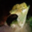}
\includegraphics[width=0.055\linewidth, bb=0 0 32 32]{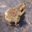}
\includegraphics[width=0.055\linewidth, bb=0 0 32 32]{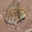}
\includegraphics[width=0.055\linewidth, bb=0 0 32 32]{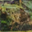}
\includegraphics[width=0.055\linewidth, bb=0 0 32 32]{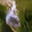}
\includegraphics[width=0.055\linewidth, bb=0 0 32 32]{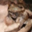}
\includegraphics[width=0.055\linewidth, bb=0 0 32 32]{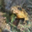}
\includegraphics[width=0.055\linewidth, bb=0 0 32 32]{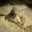}
\includegraphics[width=0.055\linewidth, bb=0 0 32 32]{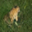}
\includegraphics[width=0.055\linewidth, bb=0 0 32 32]{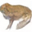}
\includegraphics[width=0.055\linewidth, bb=0 0 32 32]{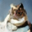}
\includegraphics[width=0.055\linewidth, bb=0 0 32 32]{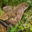}
\includegraphics[width=0.055\linewidth, bb=0 0 32 32]{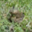}
\includegraphics[width=0.055\linewidth, bb=0 0 32 32]{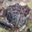}
\includegraphics[width=0.055\linewidth, bb=0 0 32 32]{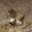}
\includegraphics[width=0.055\linewidth, bb=0 0 32 32]{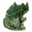}\\
\includegraphics[width=0.055\linewidth, bb=0 0 32 32]{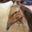}
\includegraphics[width=0.055\linewidth, bb=0 0 32 32]{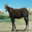}
\includegraphics[width=0.055\linewidth, bb=0 0 32 32]{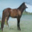}
\includegraphics[width=0.055\linewidth, bb=0 0 32 32]{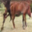}
\includegraphics[width=0.055\linewidth, bb=0 0 32 32]{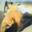}
\includegraphics[width=0.055\linewidth, bb=0 0 32 32]{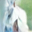}
\includegraphics[width=0.055\linewidth, bb=0 0 32 32]{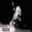}
\includegraphics[width=0.055\linewidth, bb=0 0 32 32]{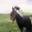}
\includegraphics[width=0.055\linewidth, bb=0 0 32 32]{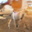}
\includegraphics[width=0.055\linewidth, bb=0 0 32 32]{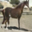}
\includegraphics[width=0.055\linewidth, bb=0 0 32 32]{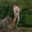}
\includegraphics[width=0.055\linewidth, bb=0 0 32 32]{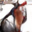}
\includegraphics[width=0.055\linewidth, bb=0 0 32 32]{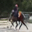}
\includegraphics[width=0.055\linewidth, bb=0 0 32 32]{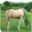}
\includegraphics[width=0.055\linewidth, bb=0 0 32 32]{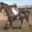}
\includegraphics[width=0.055\linewidth, bb=0 0 32 32]{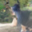}\\
\includegraphics[width=0.055\linewidth, bb=0 0 32 32]{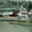}
\includegraphics[width=0.055\linewidth, bb=0 0 32 32]{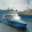}
\includegraphics[width=0.055\linewidth, bb=0 0 32 32]{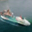}
\includegraphics[width=0.055\linewidth, bb=0 0 32 32]{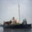}
\includegraphics[width=0.055\linewidth, bb=0 0 32 32]{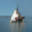}
\includegraphics[width=0.055\linewidth, bb=0 0 32 32]{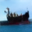}
\includegraphics[width=0.055\linewidth, bb=0 0 32 32]{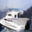}
\includegraphics[width=0.055\linewidth, bb=0 0 32 32]{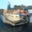}
\includegraphics[width=0.055\linewidth, bb=0 0 32 32]{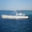}
\includegraphics[width=0.055\linewidth, bb=0 0 32 32]{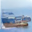}
\includegraphics[width=0.055\linewidth, bb=0 0 32 32]{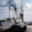}
\includegraphics[width=0.055\linewidth, bb=0 0 32 32]{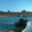}
\includegraphics[width=0.055\linewidth, bb=0 0 32 32]{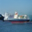}
\includegraphics[width=0.055\linewidth, bb=0 0 32 32]{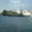}
\includegraphics[width=0.055\linewidth, bb=0 0 32 32]{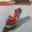}
\includegraphics[width=0.055\linewidth, bb=0 0 32 32]{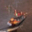}\\
\includegraphics[width=0.055\linewidth, bb=0 0 32 32]{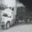}
\includegraphics[width=0.055\linewidth, bb=0 0 32 32]{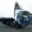}
\includegraphics[width=0.055\linewidth, bb=0 0 32 32]{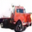}
\includegraphics[width=0.055\linewidth, bb=0 0 32 32]{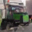}
\includegraphics[width=0.055\linewidth, bb=0 0 32 32]{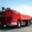}
\includegraphics[width=0.055\linewidth, bb=0 0 32 32]{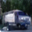}
\includegraphics[width=0.055\linewidth, bb=0 0 32 32]{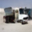}
\includegraphics[width=0.055\linewidth, bb=0 0 32 32]{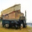}
\includegraphics[width=0.055\linewidth, bb=0 0 32 32]{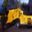}
\includegraphics[width=0.055\linewidth, bb=0 0 32 32]{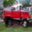}
\includegraphics[width=0.055\linewidth, bb=0 0 32 32]{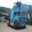}
\includegraphics[width=0.055\linewidth, bb=0 0 32 32]{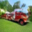}
\includegraphics[width=0.055\linewidth, bb=0 0 32 32]{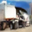}
\includegraphics[width=0.055\linewidth, bb=0 0 32 32]{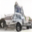}
\includegraphics[width=0.055\linewidth, bb=0 0 32 32]{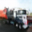}
\includegraphics[width=0.055\linewidth, bb=0 0 32 32]{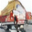}
    \par\vspace{0.5em}
    \textbf{FID=6.86149}
    \caption{Generated $32\times 32$ images of CIFAR10 \cite{cifar10} using conditional UDPM, requiring only 3 diffusion steps; equivalent to 0.3 traditional denoising step.}
    \label{fig:appndx_CIFAR10}
\end{figure*}

We provide below additional generation results of UDPM on the CIFAR10 dataset, as can be seen in Figure \ref{fig:appndx_CIFAR10}. We also provide an additional quantitative comparison to EDM \cite{karras2022elucidating} on the FFHQ \cite{karras2019style} and AFHQv2 \cite{choi2020stargan} datasets, as can be seen in Table \ref{table:appndx_FFHQ_AFHQ_FID}.

\begin{table}[!h]
    \centering
    \begin{tabular}{ | c | c  c | c  c |  }
    \hline
    \centering
                                                            & \multicolumn{2}{c|}{FFHQ} & \multicolumn{2}{c|}{AFHQv2}  \\ \hline   
                                                            & steps &    FID & steps &    FID  \\ \hline   
    EDM   \cite{karras2022elucidating}                      & 79/5  &  2.39/344.763 & 79/5  &  1.96/266.024  \\ \hline
    \textbf{UDPM (ours)}                                    & \textbf{<1}  &  \textbf{7.41} & \textbf{<1}  &  \textbf{7.10}  \\ \hline
    
    \end{tabular}
    \vspace{3mm}
\caption{FID scores comparison between UDPM and EDM \cite{karras2022elucidating} on the FFHQ \cite{karras2019style} and AFHQv2 \cite{choi2020stargan} datasets. UDPM requires 3 diffusion steps, which is equivalent to 0.3 denoising steps of EDM.}
\label{table:appndx_FFHQ_AFHQ_FID}
\end{table}

\newpage
\begin{table}[!h]
\centering
    \begin{tabular}{ | c | c | c | c | }
    \hline
    \centering
                 & CIFAR10  &  AFHQv2 & FFHQ   \\ \hline 
    Learning rate& $10^{-4}$ &  $10^{-4}$     &  $10^{-4}$     \\ \hline
    Warmup steps & 5000 &  5000     &  5000     \\ \hline
    Batch Size     & 64 &  64     &  64     \\ \hline
    Dropout     & 0.1 &  0.1     &  0    \\ \hline
    Optimizer & Adam & Adam & Adam \\ \hline
    Number of GPUS & 1 & 1 & 1\\ \hline
    \end{tabular} 
\caption{Training hyperparameters.}
\label{table:appndx_train_configs}
\end{table}

\begin{table}[!h]
\centering
    \begin{tabular}{ | c | c | c | c | }
    \hline
    \centering
                 & CIFAR10  &  AFHQv2 & FFHQ   \\ \hline 
    Architecture & NCSN \cite{song2020score}  & NCSN \cite{song2020score} & NCSN \cite{song2020score}     \\ \hline 
    Parameters & 57.73M &  65.41M     &  65.41M     \\ \hline
    Base channels & 128 &  128   &  128     \\ \hline
    Channels Multiplier     & 2, 2, 2 &  1, 2, 2, 2  &  1, 2, 2, 2     \\ \hline
    Attention resolution     & 8, 4, 2 &  16, 8, 4     &  16, 8, 4    \\ \hline
    In channels             & 3 &  3     &  3    \\ \hline
    Out channels            & 12 &  12     &  12    \\ \hline
    Blocks per scale        & 4 &  4     &  4    \\ \hline
    \end{tabular} 
\caption{Model hyperparameters.}
\label{table:appndx_Model_config}
\end{table}

\begin{table}[!h]
\centering
\resizebox{\linewidth}{!}{
    \begin{tabular}{ | c | c | c | c | }
    \hline
    \centering
                 & CIFAR10  &  AFHQv2 & FFHQ   \\ \hline 
    Architecture & DDGAN \cite{DDGAN}  & DDGAN \cite{DDGAN} & DDGAN \cite{DDGAN}     \\ \hline 
    Parameters & 18.95M &  23.86M     &  23.86M     \\ \hline
    Channels     & 192, 384, 512, 512, 512 &  192, 384, 512, 512, 512, 512  &  192, 384, 512, 512, 512, 512 \\ \hline
    In channels             & 3 &  3     &  3    \\ \hline
    Blocks per scale        & 1 &  1     &  1    \\ \hline
    Blocks type        & NCSN down &  NCSN down    &  NCSN down    \\ \hline
    Grad Penalty weight \cite{DDGAN} & 0.025, 0.1, 0.4 &  0.1, 0.4, 0.8    &  0.1, 0.4, 0.8    \\ \hline
    Dropout        & 0.4 &  0.4    &  0.3    \\ \hline
    \end{tabular} 
    }
    \caption{Discriminator network hyperparameters.}
    \label{table:appndx_Discriminator_config}
\end{table}

\end{document}